\documentclass[sigconf]{acmart}
\usepackage[ruled]{algorithm2e}
\usepackage{mathtools}
\usepackage{bm}
\usepackage{bbm}
\usepackage{amsmath}
\usepackage{multirow}
\usepackage{caption}
\usepackage{subcaption}
\usepackage{balance}
\AtBeginDocument{%
  }

\copyrightyear{2026}
\acmYear{2026}
\setcopyright{cc}
\setcctype{by}
\acmConference[WWW '26]{Proceedings of the ACM Web Conference 2026}{April 13--17, 2026}{Dubai, United Arab Emirates}
\acmBooktitle{Proceedings of the ACM Web Conference 2026 (WWW '26), April 13--17, 2026, Dubai, United Arab Emirates}
\acmPrice{}
\acmDOI{10.1145/3774904.3792322}
\acmISBN{979-8-4007-2307-0/2026/04}
\acmDOI{10.1145/3774904.3792322}

\begin{document}

\title{AHBid: An Adaptable Hierarchical Bidding Framework for Cross-Channel Advertising}

\author{Xinxin Yang}
\affiliation{%
  \institution{OPPO}
  \city{Shenzhen}
  \state{Guangdong}
  \country{China}}
\email{yangxinxin@oppo.com}

\author{Yangyang Tang}
\authornote{Corresponding author}
\affiliation{%
  \institution{OPPO}
  \city{Shenzhen}
  \state{Guangdong}
  \country{China}}
\email{tangyangyang@oppo.com}

\author{Yikun Zhou}
\affiliation{%
  \institution{OPPO}
  \city{Shenzhen}
  \state{Guangdong}
  \country{China}}
\email{zhouyikun@oppo.com}

\author{Yaolei Liu}
\affiliation{%
  \institution{OPPO}
  \city{Shenzhen}
  \state{Guangdong}
  \country{China}}
\email{liuyaolei@oppo.com}

\author{Yun Li}
\affiliation{%
  \institution{OPPO}
  \city{Shenzhen}
  \state{Guangdong}
  \country{China}}
\email{liyun3@oppo.com}

\author{Bo Yang}
\affiliation{%
  \institution{OPPO}
  \city{Shenzhen}
  \state{Guangdong}
  \country{China}}
\email{yangbo-m@oppo.com}

\renewcommand{\shortauthors}{Xinxin Yang et al.}

\begin{abstract}
  In online advertising, the inherent complexity and dynamic nature of advertising environments necessitate the use of auto-bidding services to assist advertisers in bid optimization.
  The complexity escalates in multi-channel scenarios, where effective allocation of budgets and constraints across channels with distinct behavioral patterns becomes critical for optimizing return on investment. 
  Current approaches predominantly employ either optimization-based strategies or reinforcement learning (RL) techniques. 
  However, optimization-based methods lack the flexibility to adapt to dynamic market conditions, while RL-based approaches struggle to capture essential historical dependencies and observational patterns within the constraints of Markov Decision Process (MDP) frameworks.
  To address these limitations, we propose AHBid, an \underline{\textbf{A}}daptable \underline{\textbf{H}}ierarchical \underline{\textbf{Bid}}ding framework that integrates generative planning with real-time control. 
  The framework employs a high-level generative planner utilizing diffusion models to dynamically allocate budgets and constraints through effective capture of historical context and temporal patterns. 
  We introduce a constraint enforcement mechanism to ensure compliance with specified constraints, complemented by a trajectory refinement mechanism that enhances adaptability to environmental changes through historical data utilization. 
  The system further incorporates a control-based bidding algorithm that synergistically combines historical knowledge with real-time information, significantly improving both adaptability and operational efficacy.
  Extensive experiments are conducted using both large-scale offline datasets and online A/B tests, demonstrating the effectiveness of AHBid by yielding a 13.57\% increase in overall return compared to existing baselines.
\end{abstract}





\maketitle

\section{Introduction}

\begin{figure}[t!]
  \centering
  \includegraphics[width=0.98\linewidth, trim=340 170 420 110, clip]{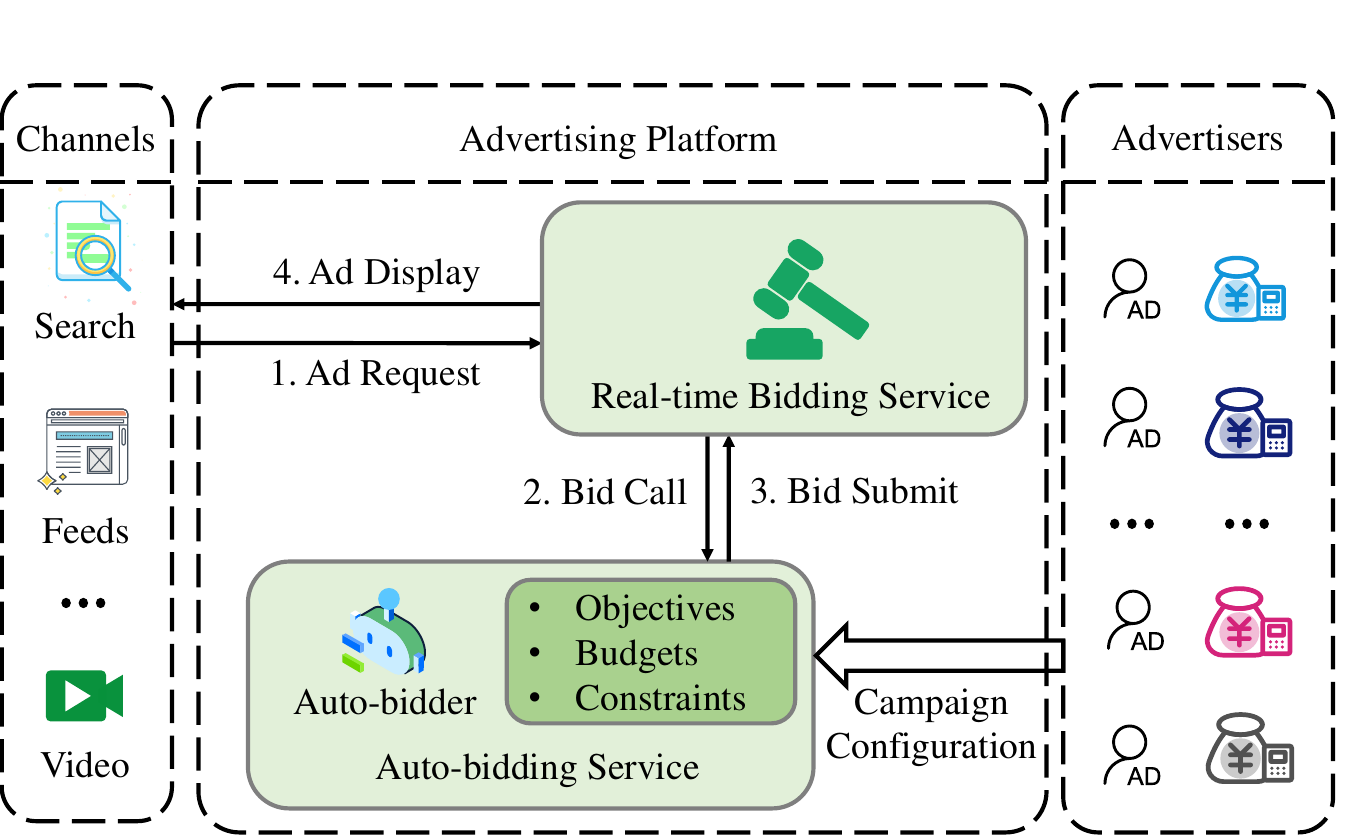}
  \caption{Illustration of auto-bidding system applications in multi-channel scenarios.}
  \label{fig:cross_channel_bidding}
  \Description{}
\end{figure}

The proliferation of real-time bidding (RTB) \cite{DBLP:conf/kdd/ChenBAD11} in programmatic advertising demands that advertisers execute bid decisions at millisecond granularity for individual advertising impressions. This transition from traditional media acquisition to impression-level auctions introduces considerable operational complexity, especially in multi-channel environments where advertisers must coordinate bidding strategies across diverse channels, such as Search, Feeds, and Video \cite{deng2023multi, DBLP:conf/www/AvadhanulaCLSS21, DBLP:journals/tc/WangTLMZDSXWW24}.
These challenges are further exacerbated by dynamic factors, such as fluctuating competitive intensity and heterogeneous user contexts.
As a result, auto-bidding services have become indispensable components of contemporary advertising systems to facilitate scalable and optimized bid management \cite{DBLP:journals/sigecom/AggarwalBBBDFGLLMMMMLPP24, DBLP:conf/cikm/OuCLDZXLTY23}.

As illustrated in Figure \ref{fig:cross_channel_bidding}, advertising platforms increasingly provide auto-bidding services across multiple channels, where an auto-bidder determines bids on behalf of advertisers for incoming impression opportunities.
Different advertising channels exhibit distinct patterns in impression volume and user behavior due to variations in user habits \cite{DBLP:journals/tc/WangTLMZDSXWW24Bak, deng2023multi, chen2025real, nuara2019dealing}. High-quality channels, characterized by elevated conversion rates (CVR) and click-through rates (CTR), generally yield superior advertising effectiveness and corresponding returns.
Strategic budget allocation across channels is essential, as inconsistent peak times and impression volumes among channels can lead to inefficient spending without careful management. By prioritizing channels better aligned with advertiser objectives and allocating budgets accordingly, advertisers can mitigate excessive expenditure in underperforming channels and enhance return on investment (ROI).
Beyond budget allocation, employing channel-specific constraints rather than uniform constraints can enhance overall effectiveness. Relaxing constraints in high-performance channels, particularly those demonstrating favorable cost-effectiveness ratios, promotes greater consumption and conversion, thereby increasing overall ROI.

Extensive research has been conducted on constrained bidding \cite{DBLP:conf/kdd/YangLWWTXG19, wu2018budget, DBLP:conf/kdd/zhao,HeCWPTYXZ21, DBLP:conf/kdd/GuanWCL0LXQXZ21}, with most studies focusing on optimizing bidding strategies within individual channels under fixed budget constraints. However, these approaches generally lack the capacity to dynamically adjust budget and constraint allocation strategies across multiple channels, thereby limiting their scalability in cross-channel bidding scenarios \cite{DBLP:journals/tc/WangTLMZDSXWW24, deng2023multi}.
In the domain of budget and constraint allocation, conventional optimization-based methods \cite{DBLP:journals/eor/LuzonPK22, DBLP:conf/www/AvadhanulaCLSS21, han2021budget} employ predictive models to estimate expected returns and derive allocation strategies through mathematical programming. Nevertheless, these bidding and allocation strategies typically operate in isolation, making them vulnerable to dynamic environmental factors that cause fluctuations in both allocation decisions and bidding behavior.
Recent reinforcement learning (RL) approaches \cite{DBLP:journals/tc/WangTLMZDSXWW24, DBLP:conf/kdd/0001HWZLYXZD25, DBLP:journals/ai/NuaraTGR22} have proposed joint optimization of bidding and allocation strategies using Markov Decision Processes (MDPs) to accommodate environmental dynamics. However, these methods face a fundamental structural limitation: the state-independence assumption inherent in the MDP framework disregards temporal dependencies and observational patterns in historical interaction sequences. This limitation impedes the identification of evolving behavioral patterns and market fluctuations, consequently undermining their practical applicability in highly volatile advertising environments.

Diffusion models \cite{sun2024conformal, DBLP:conf/iclr/WangHZ23, DBLP:conf/nips/HeBXY0WZ023, liang2023adaptdiffuser} have recently emerged as a powerful framework capable of effectively capturing temporal dependencies and historical context. Their application to allocation sequence modeling represents a promising direction for strategy enhancement, as the generative foundation of diffusion models enables explicit modeling of temporal patterns and historical bidding context, facilitating adaptive decision-making aligned with dynamic advertising environments \cite{DBLP:conf/kdd/GuoHZWYXZZ24, DBLP:conf/sigir/GaoLMJ0W0P0G0025}.
However, several challenges arise when implementing diffusion models for budget and constraint allocation tasks. First, the sequences generated by diffusion models cannot adequately guarantee constraint satisfaction, since embedded constraint information exhibits limited expressive capacity in this modeling framework. Second, the allocation results produced by diffusion models may not be fully achievable in practice, indicating insufficient adaptation to real-time environments.

To address these limitations, we propose AHBid, an \underline{\textbf{A}}daptable \underline{\textbf{H}}ierarchical \underline{\textbf{Bid}}ding framework based on diffusion models. AHBid comprises two core modules: a high-level planner responsible for allocating budgets and constraints across channels, and a low-level bidder that determines bids for individual impressions. The planner employs diffusion models to dynamically generate goal sequences, where each goal specifies allocated budget and constraint values. To mitigate constraint violation issues, we design a constraint enforcement mechanism that adaptively identifies and modulates violation loss, thereby promoting the generation of constraint-compliant trajectories. Additionally, a trajectory refinement mechanism leverages historical data to inform future goal generation, enhancing responsiveness and adaptability to environmental changes.
Given that platform-allocated budgets and constraints are likely changing over time, bidding strategies must dynamically adapt to these changing conditions. Accordingly, we introduce a control-based bidding algorithm that integrates a historical model with a real-time model, enabling enhanced adaptability and control in dynamic environments through the concurrent utilization of historical knowledge and real-time information.

The main contributions of this work are threefold:
\begin{itemize}
  \item We investigate the \textbf{c}ross-\textbf{c}hannel \textbf{c}onstrained bidding ($\text{c}^3$-bidding) problem and propose AHBid, an adaptive hierarchical bidding framework. AHBid features a generative planner capable of effectively capturing temporal dependencies and historical context, thereby enabling robust adaptation to highly volatile advertising environments.
  \item We introduce a constraint enforcement mechanism and a trajectory refinement mechanism to promote the generation of constraint-compliant trajectories while enhancing responsiveness and adaptability to environmental changes. Additionally, we design a control-based bidding algorithm that integrates historical knowledge with real-time information to improve adaptability and efficiency.
  \item We conduct extensive evaluations using large-scale simulations and real-world A/B testing, demonstrating consistent and statistically significant performance improvements across diverse bidding scenarios, confirming the framework's effectiveness and practical applicability.
\end{itemize}

\section{Hierarchical Bidding Framework}
\label{sec:hierarchical_framework}

In this section, we formalize the $\text{c}^3$-bidding problem and introduce a hierarchical framework designed to enhance auto-bidding performance through the integration of a high-level goal planner.

\subsection{Problem Formulation}
\label{subsec:problem_formulation}

We investigate the $\text{c}^3$-bidding problem, wherein an auto-bidder participates in auctions with the objective of maximizing the total value of winning impressions while adhering to a set of constraints.
Let $J$ denote the number of channels, with each channel $j$ containing $I_j$ impression opportunities.
The $i$-th impression opportunity within channel $j$ is denoted as $o_{i, j}$. 
The constraints imposed by the advertiser, namely the total budget and target cost-per-click (CPC), are represented by $B$ and $CPC$, respectively.
The auto-bidder aims to secure impressions that maximize the total impression value while complying with the constraints defined by $B$ and $CPC$.

Upon receiving impression $o_{i, j}$, the auto-bidder evaluates its associated value $v_{i, j}$ and submits a bid $b_{i, j}$.
The auto-bidder wins the impression if its submitted bid $b_{i, j}$ exceeds the highest competing bid $b^\prime_{i, j}$ from other advertisers.
If the auto-bidder successfully wins impression $o_{i, j}$, it acquires the impression's value $v_{i, j}$ and is obligated to pay the winning price $wp_{i, j}$.
We define the binary indicator $x_{i, j} \coloneqq \mathbbm{1}\{b_{i, j} > b^\prime_{i, j}\}$, which indicates the outcome of the auto-bidder's attempt to win impression $o_{i, j}$.
Formally, the $\text{c}^3$-bidding problem can be expressed as follows:
\begin{equation}
  \label{eq:bid_optimization}
  \begin{aligned}
    \max_{\bm{b}} &\sum_{j}\sum_{i} x_{i, j} \cdot v_{i, j},  \\
    \text{s.t.} \quad &\sum_{j}\sum_{i} x_{i, j} \cdot wp_{i, j} \le B,  \\
    \quad &\frac{\sum_{j}\sum_{i} x_{i, j} \cdot wp_{i, j}}{\sum_{j}\sum_{i} x_{i, j}} \le CPC ,
  \end{aligned}
\end{equation}
where $\bm{b}$ denotes the set of submitted bids.

The core challenge in addressing this problem lies in identifying impressions characterized by high values and low payment prices, i.e., impressions with favorable \textit{cost-performance ratios}, defined as $CPR_{i, j} \coloneqq \frac{v_{i, j}}{wp_{i, j}}$ for each impression $o_{i, j}$.
However, the inherent uncertainty surrounding both the values and prices of incoming impressions renders this an online stochastic knapsack problem \cite{hao2020dynamic}.
Upon the arrival of an impression $o_{i, j}$, the auto-bidder can only observe the value $v_{i, j}$ and winning price $wp_{i, j}$ after successfully acquiring it.
Therefore, it is essential for the auto-bidder to effectively estimate cost-performance ratios in advance.

Accurately capturing the distribution of cost-performance ratios presents a significant challenge for two primary reasons. First, the value and price distributions of incoming impressions exhibit considerable randomness, complicating the prediction of the cost-performance ratio for each impression. Second, the temporal distributions of impression opportunities vary substantially across channels, further increasing the complexity of the problem.

\begin{figure}[t!]
	\centering
	\begin{subfigure}{0.49\linewidth}
		\centering
		\includegraphics[width=1.0\linewidth]{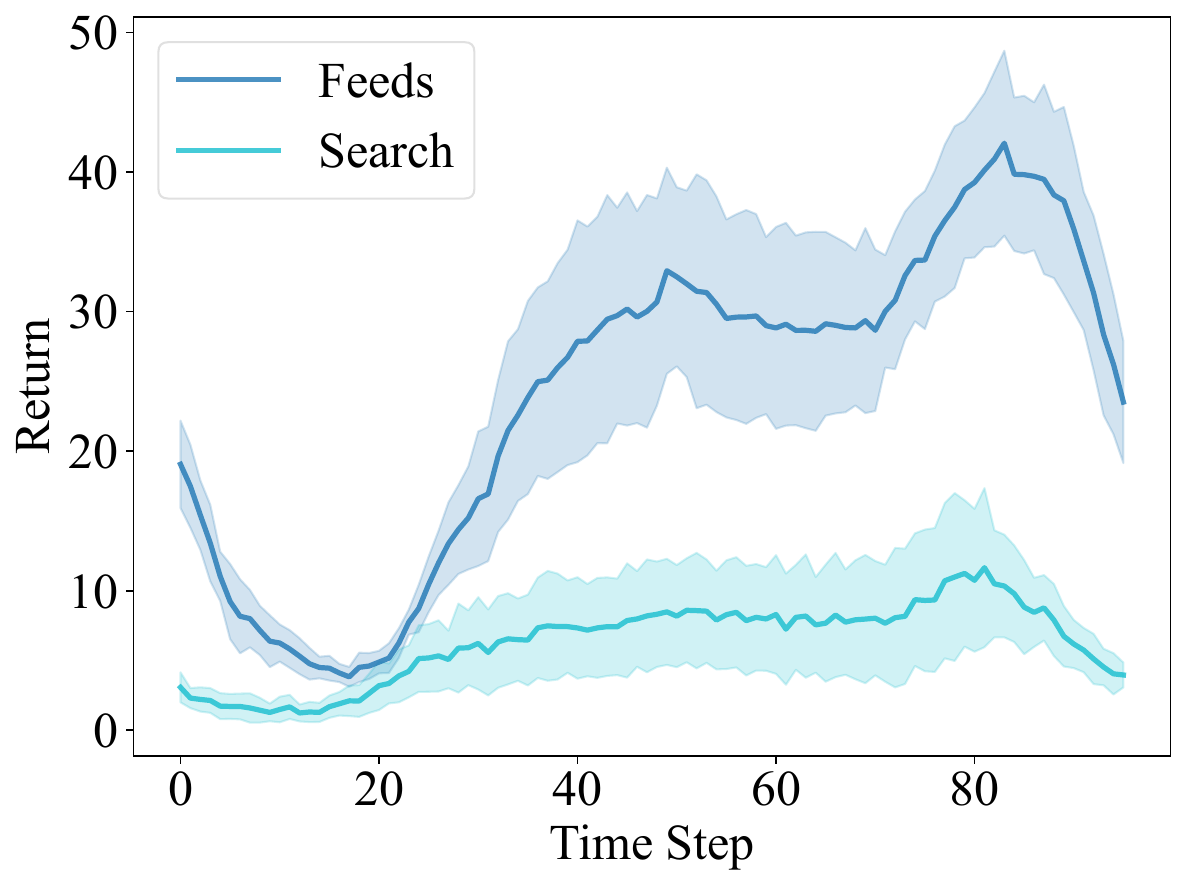}
		\caption{Comparison of Returns}
		\label{fig:return_comparison}
	\end{subfigure}
	\centering
	\begin{subfigure}{0.49\linewidth}
		\centering
		\includegraphics[width=1.0\linewidth]{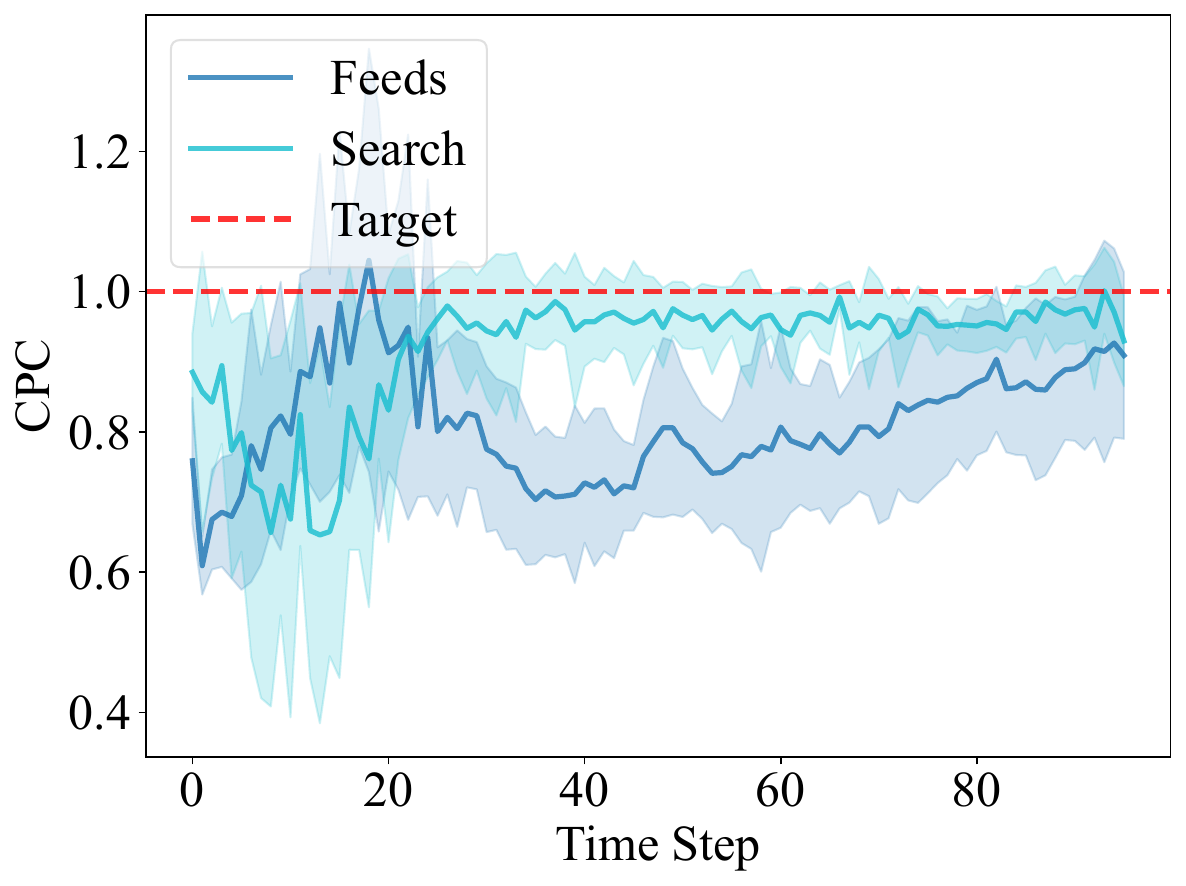}
		\caption{Comparison of CPCs}
		\label{fig:CPC_comparison}
	\end{subfigure}
  \caption{Comparative analysis of advertising performance across channels.}
	\label{fig:channel_comparison}
\end{figure}

\subsection{Hierarchical Framework}
\label{subsec:hierarchical_modeling}

To thoroughly investigate the distribution of impression opportunities, we comparatively analyze the advertising performance by deploying campaigns on both Feeds and Search, with each day partitioned into 96 consecutive 15-minute intervals. As shown in Figure \ref{fig:channel_comparison}, results indicate that Feeds yield higher returns while more consistently satisfying CPC constraints, demonstrating greater cost-effectiveness. Both channels exhibit strong periodic trends: impression volumes peak during midday and evening with increased CPC stability, while early morning intervals show reduced volume and higher volatility. These findings suggest advertisers should strategically allocate budgets toward channels and time intervals with greater CPC stability and higher impression availability. To facilitate this, it is essential to accurately capture periodic patterns, identify high-efficiency stages, and proactively allocate higher goals to these optimal periods.

Based on the aforementioned insights, we reformulate the $\text{c}^3$-bidding problem from a hierarchical perspective by augmenting the conventional auto-bidder with a high-level goal planner. The planner explicitly partitions the bidding episode for each channel into $M$ temporal stages, with each stage assigned a specific goal. This formulation yields a goal trajectory $\bm{\tau}$, defined as follows:
\begin{equation}
  \label{eq:goal_trajectory}
  \bm{\tau} \coloneqq \left[\begin{array}{llll}
    g_{1, 1} & g_{1, 2} & \ldots & g_{1, M} \\
    g_{2, 1} & g_{2, 2} & \ldots & g_{2, M} \\
    \ldots & \ldots & \ldots & \ldots \\
    g_{J, 1} & g_{J, 2} & \ldots & g_{J, M}
    \end{array}\right],
\end{equation}
where $g_{j, m}$ denotes the goal assigned to stage $m$ in channel $j$. Each goal comprises reference values, such as desired budget consumption and target CPC, which provide auxiliary guidance to the low-level auto-bidder. The auto-bidder subsequently adjusts its bidding strategy in accordance with these assigned stage-wise goals.

The optimization of the goal trajectory $\bm{\tau}$ is critical,  as it directly governs the overall efficacy of the hierarchical bidding strategy. We formalize this optimization challenge as the following problem:
\begin{equation}
  \label{eq:goal_optimization}
  \begin{aligned}
    \max_{\bm{\tau}}~& \sum_{j=1}\sum_{m=1} R_{j, m}(g_{j, m}), \\
    \text{s.t.}~&\sum_{j=1}\sum_{m=1} C_{j, m}(g_{j, m}) \le B, \\
    &\frac{\sum_{j=1}\sum_{m=1} C_{j, m}(g_{j, m})}{\sum_{j=1}\sum_{m=1} N_{j, m}(g_{j, m})} \le CPC, 
  \end{aligned}
\end{equation}
where $R_{j,m}(g_{j, m})$, $C_{j,m}(g_{j, m})$, $N_{j,m}(g_{j, m})$ denote the expected value, cost, and number of clicks, respectively, generated in stage $m$ of channel $j$ under goal $g_{j, m}$. 
Within this hierarchical framework, the auto-bidder operates by pursuing the stage-specific goals assigned by the planner, rather than optimizing the overall advertising objective. It is noteworthy that the bidder may incorporate any bidding algorithms, such as PID controllers or RL methods, to achieve these predefined stage-specific goals.

The integration of a goal planner offers three key advantages in the $\text{c}^3$-bidding context.
First, stage-aggregated statistics exhibit significantly lower volatility than impression-level data, stabilizing decision-making. Second, this enhanced stability enables more reliable estimation of cost-performance distributions, allowing strategic budget reallocation toward high-return stages. Third, the auto-bidder's decision complexity is substantially reduced by operating within shorter, stage-confined horizons rather than full-episode reasoning. This hierarchical decomposition promotes both tractability and operational efficiency.

\section{AHBid}
\label{sec:AHBid}

\begin{figure*}[t!]
  \centering
  \includegraphics[width=0.98\textwidth, trim=160 10 100 10, clip]{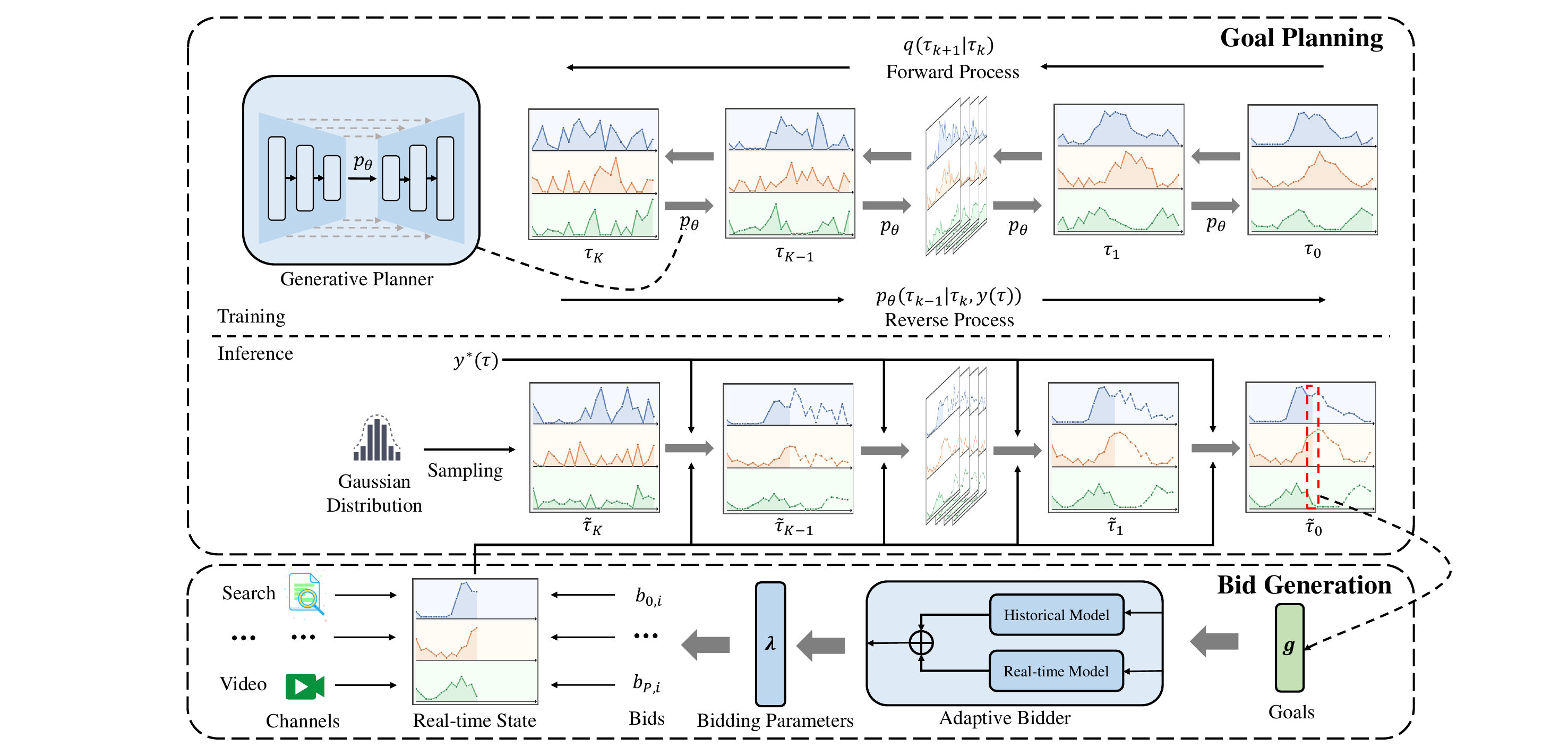}
  \caption{The overall procedure of AHBid. During training, the planner incrementally adds noise to goal trajectories and learns denoising-based reconstruction. During inference, it iteratively denoises and refines a sampled trajectory using historical data from prior stages. The optimized trajectory subsequently guides the auto-bidder in adjusting parameters and determining real-time optimal bids.}
  \Description{framework}
  \label{fig:framework}
\end{figure*}

A straightforward approach to obtaining the optimal goal trajectory involves fitting stage-specific value functions, i.e., $R_{j, m}(\cdot)$, $C_{j, m}(\cdot)$ and $N_{j, m}(\cdot)$, across all stages and solving an optimization problem. 
However, this method suffers from significant sample inefficiency and limited adaptability to dynamic environments, as it requires extensive offline data from all channels to accurately parameterize these functions. Alternative RL-based approaches struggle to identify evolving behavioral patterns and market fluctuations within the constraints of MDP frameworks, leading to suboptimal performance in highly volatile advertising environments.

To overcome these limitations, we propose AHBid, a framework featuring an adaptable generative planner. This planner employs diffusion models to produce goal trajectories and adaptively refines them using historical data, thus enhancing adaptability to environmental changes. Subsequently, an adaptive bidder optimizes real-time bidding parameters to decide bids for incoming impression opportunities. The overall training and inference workflow is illustrated in Figure \ref{fig:framework}.

\subsection{Framework Overview}
\label{subsec:overview}

The optimization of the goal trajectory is critical, as its quality directly determines the overall efficacy of the hierarchical bidding strategy. While optimization-based methods and RL approaches suffer from sample inefficiency and limited adaptability, we introduce an adaptable generative planner based on diffusion models.
Specifically, the planner jointly models the distribution of goal trajectories $\bm{\tau}$ and their associated properties $\bm{y}(\bm{\tau})$, generating trajectories via maximum likelihood estimation (MLE):
\begin{equation}
  \label{eq:mle}
  \begin{aligned}
    \max_{\theta} \mathbb{E}_{\bm{\tau}\sim D}\left[\log p_{\theta}\left(\bm{\tau}|\bm{y}(\bm{\tau})\right)\right],
  \end{aligned}
\end{equation}
where $\bm{y}(\bm{\tau})$ denotes properties of the trajectory $\bm{\tau}$, such as budget, constraints and target metrics including total return. By maximizing this objective, the planner can sample trajectories $\bm{\tau}$ that align with the conditioning information $\bm{y}(\bm{\tau})$. 

To facilitate both training and inference, we preprocess the dataset through aggregation and scaling operations. Specifically, we apply the mini-batch k-means++ algorithm \cite{sculley2010web} to cluster trajectories exhibiting similar constraints and advertising objectives, forming multiple sub-datasets. Within each sub-dataset, trajectory returns are normalized as follows:
\begin{equation}
  \label{eq:normalize_return}
  \begin{aligned}
    R = \frac{R_{\bm{\tau}} - R_{min}}{R_{max}-R_{min}},
  \end{aligned}
\end{equation}
where $R_{\bm{\tau}}$ denotes the return of trajectory $\bm{\tau}$, $R_{max}$ and $R_{min}$ represent the maximum and minimum returns within the sub-dataset, respectively. This normalization rescales returns into the interval $[0, 1]$, which is then incorporated into $\bm{y}(\bm{\tau})$. Note that trajectories with higher returns yield larger normalized values; hence, $R=1$ corresponds to the optimal trajectory with the highest returns. During generation, we set $R=1$ to produce a trajectory aligned with the maximum return condition, thereby optimizing advertiser returns.

Upon receiving high-level goal trajectories, the low-level bidder must generate bids for incoming impressions to achieve these predefined goals, which is a problem analogous to Eq. (\ref{eq:bid_optimization}). However, determining bids individually for each impression is challenging due to the vast number of impressions and the inherent unpredictability of their performance prior to arrival. Fortunately, prior works \cite{HeCWPTYXZ21, DBLP:conf/kdd/YangLWWTXG19} have derived a unified optimal bidding function using primal-dual theory, which reduces the solution space dimensionality to the number of constraints. The optimal bid is expressed as:
\begin{align}
  b^* = \lambda_0 \cdot v + \lambda_1 \cdot CPC \cdot CTR,
  \label{eq:optimal_bidding_solution}
  \end{align}
where $v$, $CPC$ and $CTR$ denote the impression value, the CPC constraint, and the click-through rate, respectively, while $\lambda_0$ and $\lambda_1$ are dual parameters that modulate bid amounts. This formulation allows the constrained bidding problem to be addressed by optimizing dual parameters rather than searching for individual bids. 

\subsection{Learning the Planner}
\label{subsec:planner_training}

Using the preprocessed trajectories, the planner employs conditional diffusion modeling to learn the distribution of goal trajectories. The modeling framework consists of two fundamental processes: the forward process and the reverse process, formally defined as:
\begin{equation}
  \label{eq:forward_backward}
  \begin{aligned}
      q(\bm{\tau}_{k+1}| \bm{\tau}_{k}), \quad p_{\theta}(\bm{\tau}_{k-1}| \bm{\tau}_{k}, \bm{y}(\bm{\tau})),
  \end{aligned}
\end{equation}
where $q(\cdot)$ denotes the forward process that incrementally adds Gaussian noise to the trajectory, while $p_{\theta}(\cdot)$ represents the reverse process in which a parameterized model $\theta$ performs iterative denoising. $\bm{\tau}_k$ represents the noised trajectory at diffusion step $k \in [1, K]$ where $K$ denotes the total number of diffusion steps.

In the forward process, the diffusion procedure is modeled as a Markov chain, where the noised trajectory $\bm{\tau}_{k}$ depends conditionally only on $\bm{\tau}_{k-1}$:
\begin{equation}
  \label{eq:forward}
  \begin{aligned}
      q(\bm{\tau}_{k}| \bm{\tau}_{k-1}) = \mathcal{N}\left(\bm{\tau}_{k}; \sqrt{1-\beta_{k}} \bm{\tau}_{k-1}, \beta_k \bm{\epsilon}\right),
  \end{aligned}
\end{equation}
where $\bm{\epsilon}$ represents sampled Gaussian noise, $\beta_k \in [0, 1]$ for $ k = 1, \ldots, K$ constitute the variance schedule governing the noise intensity. To ensure a smooth noise progression, we adopt a cosine schedule \cite{nichol2021improved} for designing $\beta_k$. As $k\rightarrow \infty$, $\bm{\tau}_k$ converges to a trajectory following a standard Gaussian distribution.

Upon obtaining the noised trajectory, the planner optimizes the model parameters $\theta$ to reconstruct the original trajectory $\bm{\tau}_0$ from $\bm{\tau}_k$ via an iterative denoising process, expressed as:
\begin{equation}
  \label{eq:sample_process}
  \begin{aligned}
      \widetilde{\bm{\tau}}_{k-1}\sim \mathcal{N}\left(\widetilde{\bm{\tau}}_{k-1}| \bm{\mu}_\theta\left(\bm{\tau}_k, \bm{y}(\bm{\tau}), k\right), \bm{\Sigma}_\theta(\bm{\tau}_{k}, k)\right),
  \end{aligned}
\end{equation}
where $\widetilde{\bm{\tau}}_{k-1}$ is the estimated denoised trajectory at step $k-1$, $\bm{\mu}_\theta(\cdot)$ and $\bm{\Sigma}_\theta(\cdot)$ represent the predicted mean and variance of the denoised trajectory, respectively. Empirical studies suggest that predicting the noise $\hat{\bm{\epsilon}}_{k}$ rather than directly estimating the mean $\bm{\mu}$ leads to lower training loss and improved sample quality \cite{DBLP:conf/nips/HoJA20}. The mean and variance are parameterized as:
\begin{align}
  \bm{\mu}_\theta\left(\bm{\tau}_k, \bm{y}(\bm{\tau}), k\right) &= \frac{1}{\sqrt{\alpha_k}}\left(\bm{\tau}_k - \frac{\beta_k}{\sqrt{1-\overline{\alpha}_k }}\hat{\bm{\epsilon}}_k\right), \label{eq:parameter_mu} \\
  \bm{\Sigma}_\theta(\cdot) &= \beta_k, \label{eq:parameter_sigma}
\end{align}
where $\alpha_k = 1-\beta_k$ and $\overline{\alpha}_k = \prod_{i=1}^k \alpha_k$.  The planner is trained by minimizing the diffusion loss:
\begin{equation}
  \label{eq:loss_diffusion}
  \begin{aligned}
      \mathcal{L}_{diff} = \mathbb{E}_{k, \bm{\tau}, \bm{\epsilon}}\left[\|  \bm{\epsilon} - \hat{\bm{\epsilon}}_k\|^2 \right].
  \end{aligned}
\end{equation}

Following \cite{DBLP:conf/iclr/AjayDGTJA23, ho2022classifier}, we employ a classifier-free guidance strategy to steer the conditional generation of goal trajectories. Specifically, we jointly train a conditional model $\bm{\epsilon}_\theta\left(\bm{\tau}_k, \bm{y}(\bm{\tau}), k\right)$ and an unconditional model $\bm{\epsilon}_\theta(\bm{\tau}_k, \varnothing, k)$ by randomly replacing the condition $\bm{y}(\tau)$ with a dummy value $\varnothing$. The final noise estimate is derived as a linear combination of the conditional and unconditional outputs:
\begin{equation}
  \label{eq:noise_estimation}
  \begin{aligned}
      \hat{\bm{\epsilon}}_k \coloneqq \bm{\epsilon}_\theta\left(\bm{\tau}_k, \varnothing, k\right) + \omega \left(\bm{\epsilon}_\theta(\bm{\tau}_k, \bm{y}(\bm{\tau}), k) - \bm{\epsilon}_\theta(\bm{\tau}_k, \varnothing, k)\right),
  \end{aligned}
\end{equation}
where $\omega$ controls the guidance strength, regulating the influence of the conditioning information on the generated output.

Despite the efficacy of the generative approach in generating suitable goal trajectories, several practical challenges remain. First, the actual state attained by the bidder may diverge from the predetermined goal, indicating that pregenerated trajectories no longer adapt adequately to the current environment. Second, the constraint information embedded within $\bm{y}(\bm{\tau})$ possesses limited expressive capacity, complicating the enforcement of strict constraint satisfaction. These issues necessitate the design of enhanced refinement and control mechanisms.

Due to the inherent randomness and dynamic nature of advertising environments, the outcomes achieved by bidders often deviate from predefined goals, indicating that pre-generated goals may no longer suit the current context and risk performance degradation.
To address this, we propose regenerating the goal plan after each stage. We further introduce an inpainting technique that leverages historical state sequences to inform future goal generation. After $T$ stages, the planner acquires the state history $\bm{s}_{1:J, 1:T}$ to refine subsequent goals. At each denoising step $k$, the trajectory $\widetilde{\bm{\tau}}_k$ is updated by integrating these historical states as:
\begin{equation}
  \label{eq:inpainting}
  \widetilde{\bm{\tau}}_k=\left[\begin{array}{llllll}
    s_{1, 1} & \ldots & s_{1, T} & \widetilde{g}_{1, T+1} & \ldots & \widetilde{g}_{1, M} \\
    s_{2, 1} & \ldots & s_{2, T} & \widetilde{g}_{2, T+1} & \ldots & \widetilde{g}_{2, M} \\
    \ldots & \ldots & \ldots & \ldots & \ldots & \ldots \\
    s_{J, 1} & \ldots & s_{J, T} & \widetilde{g}_{J, T+1} & \ldots & \widetilde{g}_{J, M}
    \end{array}\right]_k,
\end{equation}
where $s_{j, t}$ denotes the actual state achieved at stage $t \in [1, T]$, and $\widetilde{g}_{j, m}$ denotes the estimated goal for future stages $m \in [T+1, M]$. By integrating historical data into the denoising process, the model adaptively adjusts future goals to align with observed patterns, thereby enhancing responsiveness and adaptability to environmental changes.

\subsubsection{Constraint Enforcement}

Incorporating constraint information directly into $\bm{y}(\bm{\tau})$ exhibits limited expressive capacity and cannot guarantee strict constraint satisfaction. A direct strategy involves employing a constraint violation loss to incentivize the model to generate constraint-satisfying trajectories.
However, evaluating constraint satisfaction during diffusion model training presents two significant challenges. First, as diffusion models predict inter-step noise rather than trajectories, directly assessing constraint violations is inherently complex. Second, intermediate diffusion steps naturally permit violations due to the inherent presence of stochastic noise.
To address these issues, we design a hybrid loss function that adaptively identifies and modulates violation loss contributions. 

We first define a constraint violation function that quantifies the overall severity of violations as follows:
\begin{align}
  V\left(\bm{\tau}_k, \bm{y}(\bm{\tau})\right) &= \kappa^b V^{B}\left(\bm{\tau}_k, \bm{y}(\bm{\tau})\right) + \kappa^c V^{C}\left( \bm{\tau}_k, \bm{y}(\bm{\tau}) \right)
  \label{equ:violation_value},
  \end{align}
where $V^{B}(\cdot)$ and $V^{C}(\cdot)$ evaluate violations of budget and CPC constraints, respectively, and $\kappa^b$ and $\kappa^c$ are weighting coefficients. The violation loss is then defined as:
\begin{align}
  \mathcal{L}_{vio}(\widetilde{\bm{\tau}}_{k}, \bm{y}(\bm{\tau})) = \mathbb{E}\left[V\left(\widetilde{\bm{\tau}}_{k}, \bm{y}(\bm{\tau})\right)\right].
  \label{equ:violation_loss}
  \end{align}
Here $\widetilde{\bm{\tau}}_{k}$ denotes the predicted denoised trajectory at diffusion step $k$, which can be computed using the estimated noise $\hat{\bm{\epsilon}}$ via Eq. (\ref{eq:sample_process}) and Eq. (\ref{eq:parameter_mu}). It should be noted that the violation strength of $\widetilde{\bm{\tau}}_{k}$ is not expected to be zero, as even the ground-truth trajectory $\bm{\tau}_{k}$ may also exhibit constraint violations due to added noise.

Fortunately, the ground-truth violation strength $V_{GT}$ for $\bm{\tau}_{k}$ can be computed by drawing $N$ samples and calculating the average violation value:
\begin{align}
  V_{GT}(\bm{\tau}_{k}, \bm{y}(\bm{\tau})) = \mathbb{E}\left[ V\left( \bm{\tau}_{k}, \bm{y}(\bm{\tau}) \right) \right].
  \label{equ:violation_groundtruth}
  \end{align}
By normalizing the violation loss with respect to this ground-truth value, we formulate the final hybrid loss function:
\begin{align}
  \mathcal{L}_{final} = \mathcal{L}_{diff} + \xi \cdot\frac{\mathcal{L}_{vio}}{V_{GT}},
  \label{eq:loss_function_hybrid}
  \end{align}
where $\xi$ is a scaling factor. This normalization ensures that when ${V}_{GT}$ is large, typically at larger diffusion steps where data is noisier, the influence of $\mathcal{L}_{vio}$ on training is proportionally reduced. Consequently, the model adaptively adjusts parameters to minimize constraint violations while preserving the quality of the generated trajectories.

\subsection{Learning the Bidder}
\label{subsec:learning_bidder}

Upon receiving a goal from the planner, the bidder must adjust the dual parameters $\lambda_0$ and $\lambda_1$ to achieve the assigned goals, which is challenging due to environmental stochasticity and uncertainty. Previous studies \cite{DBLP:conf/kdd/GuoHZWYXZZ24, DBLP:conf/icml/LiW0Z23} have employed inverse dynamics models for parameter prediction. However, these models are typically trained on static offline datasets and exhibit limited adaptability, reducing their efficacy in dynamic real-time bidding environments.
To overcome these limitations, we propose an adaptive bidding strategy that integrates both historical data and real-time information. This approach employs two complementary models: a historical model and a real-time model, which work collaboratively to determine optimal bid prices.

The historical model employs model predictive control (MPC) \cite{kouvaritakis2016model} to capture the coupling effect of dual variables on budget expenditure and CPC. Given the impracticality of fully modeling the highly nonlinear advertising environment, we approximate this coupling using a linear representation:
\begin{align}
  \left[
  \begin{matrix}
  Cost \\
  CPC \\
  \end{matrix}
  \right] = 
  \left[
  \begin{matrix}
  \bm{x} \quad \bm{b}
  \end{matrix}
  \right] 
  \left[
    \begin{matrix}
    \lambda_0 \\
    \lambda_1 \\
    1 \\
    \end{matrix}
    \right],
\label{equ:system_dynamic}
\end{align}
where $\bm{x}$ is a $2\times 2$ matrix and $\bm{b}$ is a $2\times 1$ vector. Given desired adjustments $\Delta Cost$ and $\Delta CPC$ from feedback, the corresponding changes in dual variables are obtained by solving:
\begin{align}\left[
  \begin{matrix}
  \Delta \lambda_0 \\
  \Delta \lambda_1 \\
  \end{matrix}
  \right] = 
  \left[
  \begin{matrix}
  \bm{x}
  \end{matrix}
  \right]^{-1}
  \left[
  \begin{matrix}
  \Delta Cost \\
  \Delta CPC \\
  \end{matrix}
  \right].
\label{equ:system_dynamic_inverse_derivation}
\end{align}
This formulation implies that adjustments to $\lambda_0$ and $\lambda_1$ are linear combinations of $\Delta Cost$ and $\Delta CPC$. The inverse transformation $\left[\bm{x}\right]^{-1}$ can be parameterized by two coefficients $\phi$ and $\varphi$:
\begin{align}
  \left[
  \begin{matrix}
  \bm{x}
  \end{matrix}
  \right]^{-1} = 
  \left[
  \begin{matrix}
  \phi \quad 1-\phi \\
  1-\varphi \quad \varphi \\
  \end{matrix}
  \right].
\end{align}
These parameters are optimized by searching for the best-fit values using historical data.

For the real-time model, we employ periodic linear programming (LP) \cite{agrawal2014dynamic} to refine dual variables using newly acquired information. Specifically, at time step $t$, the model aggregates impression opportunities over a sliding window $[t-T_0, t]$, where $T_0$ denotes the observation length. The collected opportunities, represented as $U(t, T_0)$, are utilized to compute optimal dual variables governing the subsequent bidding period. This method relies on the assumptions of short-term environmental stationarity and consistency in impression distribution across consecutive intervals.

Finally, the outputs of the historical and real-time models, denoted as $\Lambda_h(t)$ and $\Lambda_r(t)$ respectively, are combined as follows:
\begin{align}
\Lambda(t) = \sigma (t) \cdot \Lambda_h(t) + (1-\sigma(t))\cdot \Lambda_r(t), \label{eq:model_combination}
\end{align}
where $\sigma(t) = 1-\min(\rho t, p)$ is a weighting factor, with $p\in [0, 1]$ and $\rho$ a scaling coefficient. The weight factor $\sigma(t)$ is initially close to 1 and gradually decreases to $1-p$ as $t$ increases. This design reflects the intuition that greater reliance is placed on the historical model during initial stages to mitigate cold-start issues, while increasing weight is assigned to the real-time model as the advertising environment stabilizes over time.

\subsection{Training and Inference}
\label{subsec:complexity_analysis}

During training, we employ the well-established DDPM \cite{DBLP:conf/nips/HoJA20} framework to train the planner. In each iteration, a set of trajectories $\bm{\tau}$ and their associated properties $\bm{y}(\bm{\tau})$ are sampled. With probability $p_u$, the condition $\bm{y}(\bm{\tau})$ is replaced with a dummy value $\varnothing$. The noised trajectory $\bm{\tau}_{k}$ is then obtained using Eq. (\ref{eq:forward}), and the corresponding noise $\hat{\bm{\epsilon}}$ is estimated using Eq. (\ref{eq:noise_estimation}). The planner is updated with the loss computed according to the hybrid objective defined in Eq. (\ref{eq:loss_function_hybrid}). For the low-level bidder, the optimization process involves directly searching for the optimal parameters of the historical model. This procedure is substantially more efficient than training the planner, owing to the fact that the historical model comprises only two parameters.

During inference, the planner makes decisions exclusively at the beginning of each stage, resulting in a total of $M$ decisions per episode. In contrast, the bidder determines bids for every individual incoming impression. The computational complexity of the planner for a single episode is $O(MKH)$, where $M$, $K$, and $H$ represent the number of stages, diffusion steps, and hidden units in the model, respectively. The complete training and inference procedure for AHBid is summarized in Appendix \ref{appendix:training_inference_procedure}.

\section{Experiments}
\label{sec:experiment}

\subsection{Experimental Setup}

\subsubsection{Dataset.}

To evaluate the performance of AHBid, we conducted offline experiments using two distinct datasets: a publicly available AuctionNet dataset\footnote{\url{https://github.com/alimama-tech/AuctionNet}} and a proprietary industrial dataset, GenB-4c. The AuctionNet dataset comprises 40,000 advertising state trajectories from advertisers on the Taobao platform, each capturing the bidding process across 500,000 impression opportunities. The GenB-4c dataset, collected from a production advertising system, contains 930,000 trajectories spanning four channels (Pop-Up, Feeds, Video, and Search) during the period from March 3 to March 30, 2025. Each trajectory in GenB-4c encompasses over 970,000 impression opportunities.

\begin{table*}[t!]
  \caption{Performance comparison with baselines.}
  \centering
  \begin{tabular}{ c|c|c c c c c c c c }
  \toprule
  \textbf{Datasets} & \textbf{Metrics} & \textbf{PID} & \textbf{USCB} & \textbf{CQL} & \textbf{ABPlanner} & \textbf{DiffBid} & \textbf{HiBid} & \textbf{AHBid} & \textit{improv} \\ \midrule
  \multirow{4}{*}{AuctionNet} & $R$ & 206.37 & 250.30 & 245.68 & 275.33 & \underline{278.16} & 250.59 & \textbf{280.94} & 0.99\% \\ 
                         & $CSR$ & $\underline{0.904}$ & 0.813 & 0.835 & 0.816 & 0.882 & 0.842 & \textbf{0.917} & 1.04\% \\
                         & $BCR$ & 0.764 & 0.837 & 0.818 & 0.879 & $\underline{0.927}$ & 0.863 & \textbf{0.936} & 0.97\% \\  \midrule
  \multirow{4}{*}{GenB-4c} & $R$ & 210.43 & 236.68 & 242.18 & 248.45 & 251.76 & \underline{275.40} & \textbf{298.59} & 8.42\% \\ 
                        & $CSR$ & \underline{0.890} & 0.798 & 0.819 & 0.799 & 0.826 & {0.856} & \textbf{0.907} & 1.91\% \\
                        & $BCR$ & 0.787 & 0.802 & 0.813 & 0.852 & 0.852 & \underline{0.869} & \textbf{0.936} & 7.71\% \\ 
  \bottomrule
  \end{tabular}
  \label{tab:performance_evaluation}
  \end{table*}

\subsubsection{Offline Evaluation System.}

We developed a Virtual Advertising System (VAS) using historical data collected from real-world advertising platforms. The VAS consists of two core components: an advertising system simulator and a user feedback predictor. The simulator replicates key operational processes, including ad retrieval, bidding, ranking, and pricing, while the predictor estimates user responses to specific advertisements to facilitate performance evaluation.
The advertising process is structured over a daily cycle, discretized into 48 time stages, each corresponding to a 30-minute interval. The volume of impression opportunities per stage varies dynamically, typically ranging from 3,000 to 25,000.

\subsubsection{Evaluation Metrics.}

The objective of advertisers is to maximize the number of conversions while adhering to budget and CPC constraints. To assess the performance of advertising campaigns, three key metrics are introduced:
\begin{itemize}
  \item {$R$}: denotes the average returns of campaigns, i.e., the amount of conversions.
  \item {$CSR$}: is the proportion of campaigns satisfying the CPC constraint (an overshooting within 10\% is acceptable).
  \item {$BCR$}: is the average budget consumption rate of campaigns. Since campaigns are suspended upon budget exhaustion, $BCR$ is always less than 1.0.
  \end{itemize}
To eliminate external influence, each model undergoes random initialization 30 times, and the average of top-5 scores is reported as the final score.

\subsubsection{Implementation Details}
\label{subsubsec:implementtation_detail}

We employ the widely adopted U-Net architecture \cite{ronneberger2015u} to implement the goal planner. The trajectory length $M$ is set to 48. The guidance scale $w$ and the conditional drop rate $p_u$ are configured as 0.2 and 0.25, respectively. The number of diffusion steps $K$ is set to 30. For optimization, we use the Adam optimizer \cite{kingma2014adam} with a batch size of 256 and a learning rate of $1e-4$. The model is trained for a total of 1,000 epochs. 

\subsection{Performance Evaluation}

We compare AHBid with six baselines:
\begin{itemize}
\item {PID} \cite{DBLP:conf/kdd/YangLWWTXG19}: is a classical control-theoretic method that maintains CPC values near target constraints to ensure performance bound adherence.
\item {USCB} \cite{HeCWPTYXZ21}: is a predecessor that formulates optimal bidding under a general theoretical framework and adaptively adjusts bids using a RL algorithm.
\item {MCQ} \cite{DBLP:conf/nips/LyuMLL22}: is an enhanced offline RL method built upon Conservative Q-Learning (CQL) \cite{DBLP:conf/nips/KumarZTL20}, which learns effective bidding policies directly from logged historical data.
\item {ABPlanner}: \cite{DBLP:conf/kdd/0001HWZLYXZD25} is an adaptive budget planner that dynamically adjusts allocations using few-shot online interaction data.
\item {DiffBid} \cite{DBLP:conf/kdd/GuoHZWYXZZ24}: utilizes a diffusion model for state trajectory generation and an inverse dynamics model for bid inference, providing a generative approach to bidding optimization.
\item {HiBid} \cite{DBLP:journals/tc/WangTLMZDSXWW24}: is a hierarchical RL framework that jointly optimizes cross-channel budget allocation and bid pricing.
\end{itemize}

The performance evaluation results are summarized in Table \ref{tab:performance_evaluation}. PID ranks second in constraint satisfaction rate, yet underperforms in both return and budget consumption rate, due to its narrow focus on CPC constraint adherence without optimizing overall returns.
Generative models like DiffBid consistently outperform RL-based methods, e.g., USCB and MCQ, as they model the joint distribution of full trajectories and returns, reducing error propagation common in RL paradigms.
Although both DiffBid and AHBid use diffusion models, AHBid achieves superior results across all datasets by integrating historical knowledge with real-time bidding information, while DiffBid relies solely on offline-trained inverse dynamic models.
Performance varies across datasets; for example, DiffBid outperforms HiBid on AuctionNet but is surpassed on GenB-4c, underscoring the need for adaptive budget allocation in cross-channel scenarios.
AHBid consistently ranks as the top-performing method in all evaluated settings. A detailed comparative analysis of the bidding model is provided in Appendix \ref{subsec:ablation_study}.

\subsection{In-depth Analysis}

\begin{figure}[t!]
	\centering
	\begin{subfigure}{0.49\linewidth}
		\centering
		\includegraphics[width=1.0\linewidth]{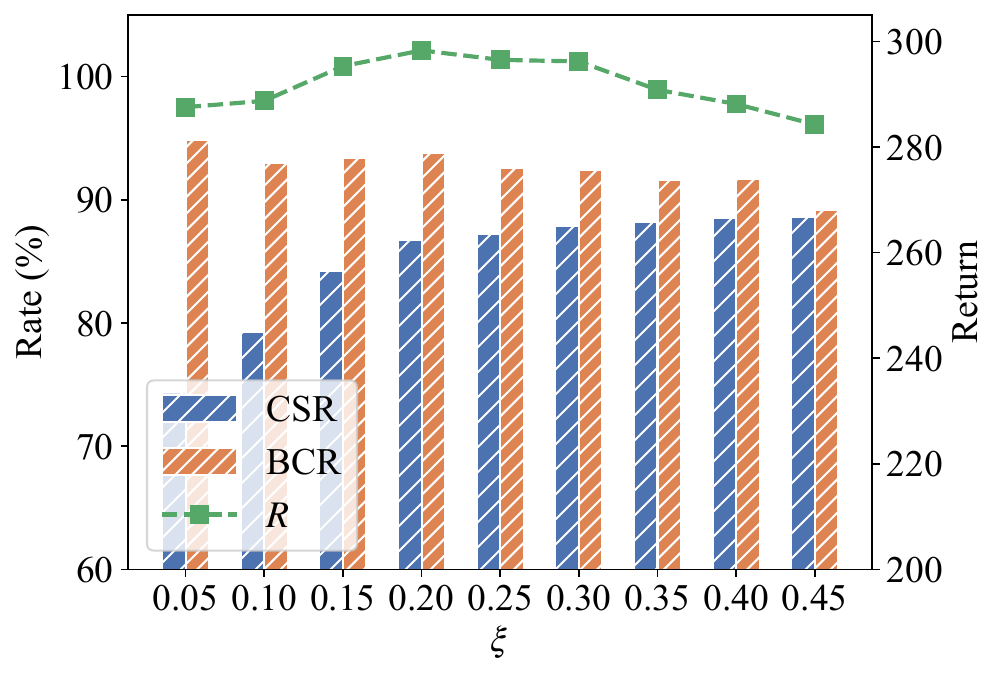}
		\caption{Impact of $\xi$}
		\label{fig:loss_weight_tuning}
	\end{subfigure}
	\centering
	\begin{subfigure}{0.49\linewidth}
		\centering
		\includegraphics[width=1.0\linewidth]{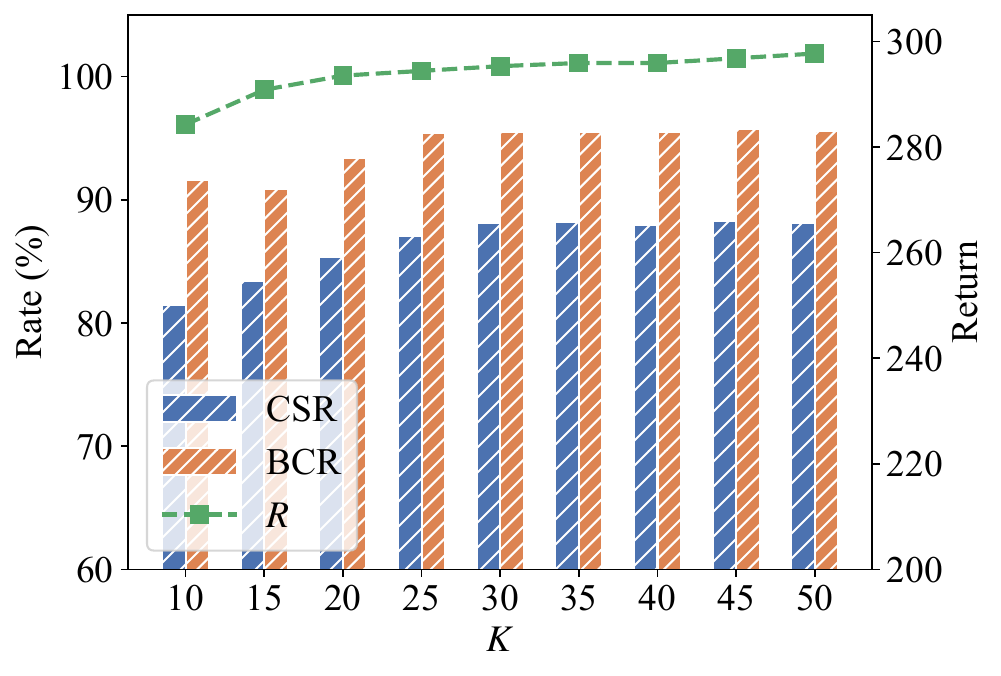}
		\caption{Impact of $K$}
		\label{fig:K_tuning}
	\end{subfigure}
  \caption{Hyper-parameter tuning.}
	\label{fig:hyper-parameter_tuning}
\end{figure}

\subsubsection{Hyper-parameter Tuning}
\label{subsec:hyper-parameter_tuning}

To investigate the effects of the violation loss factor $\xi$ and the number of diffusion steps $K$, we conducted a comprehensive hyperparameter analysis. The values of $\xi$ and $K$ are varied within [0.05, 0.45] and [10, 50], respectively.

As shown in Figure \ref{fig:loss_weight_tuning}, the constraint satisfaction rate increases with higher $\xi$, while both budget spend rate and return decline beyond a threshold. This indicates that larger $\xi$ prioritizes constraint adherence but may impair overall returns when excessive.
We further evaluated the impact of $K$, a critical parameter influencing both computational efficiency and model performance. Results in Figure \ref{fig:K_tuning} show that performance improves with increasing $K$ but exhibits diminishing returns for $K>30$. Thus, $K=30$ is selected as the optimal configuration for AHBid.

\subsubsection{Study of State Transition and Distribution}

\begin{figure}[t!]
	\centering
	\begin{subfigure}{0.49\linewidth}
		\centering
		\includegraphics[width=1.0\linewidth]{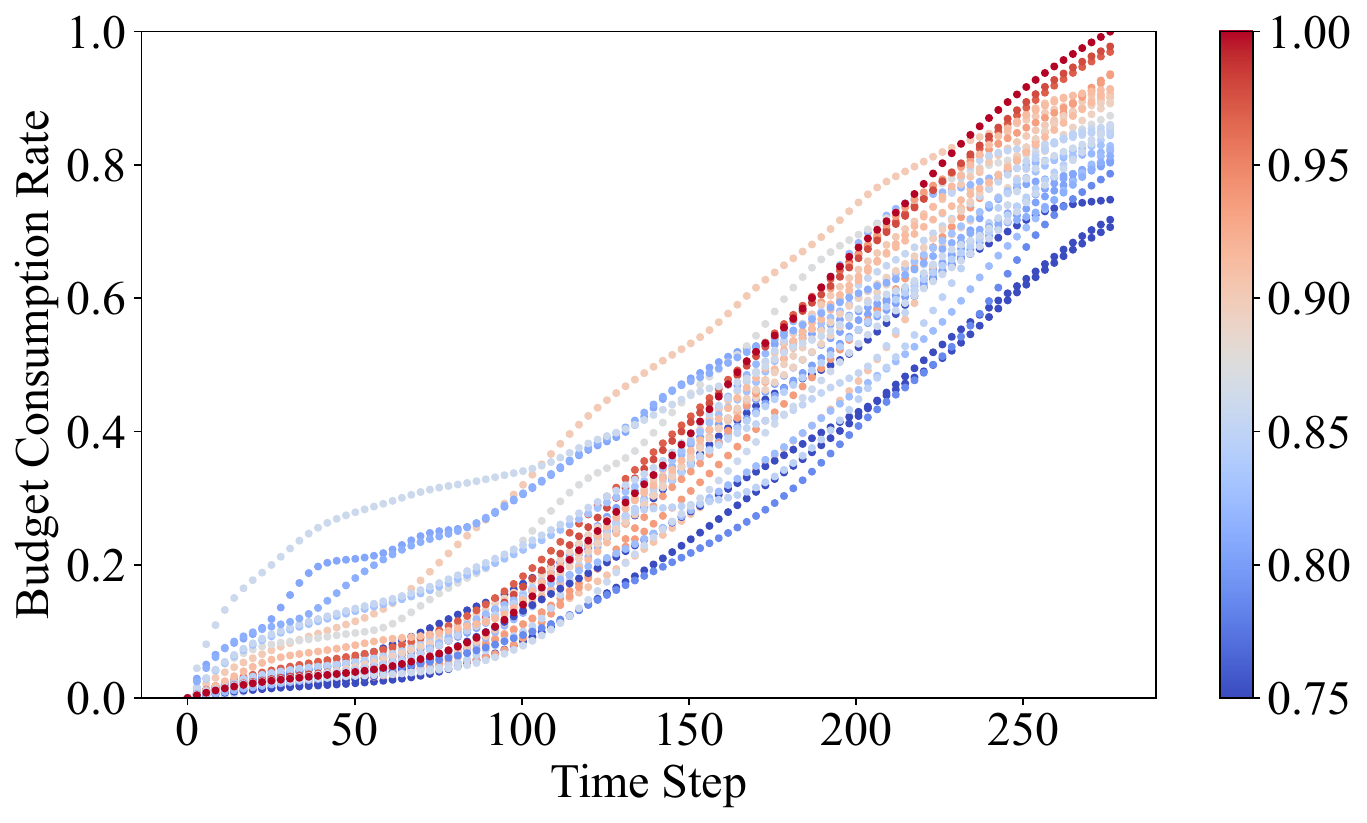}
		\caption{HiBid}
		\label{fig:cost_hibid}
	\end{subfigure}
	\centering
	\begin{subfigure}{0.49\linewidth}
		\centering
		\includegraphics[width=1.0\linewidth]{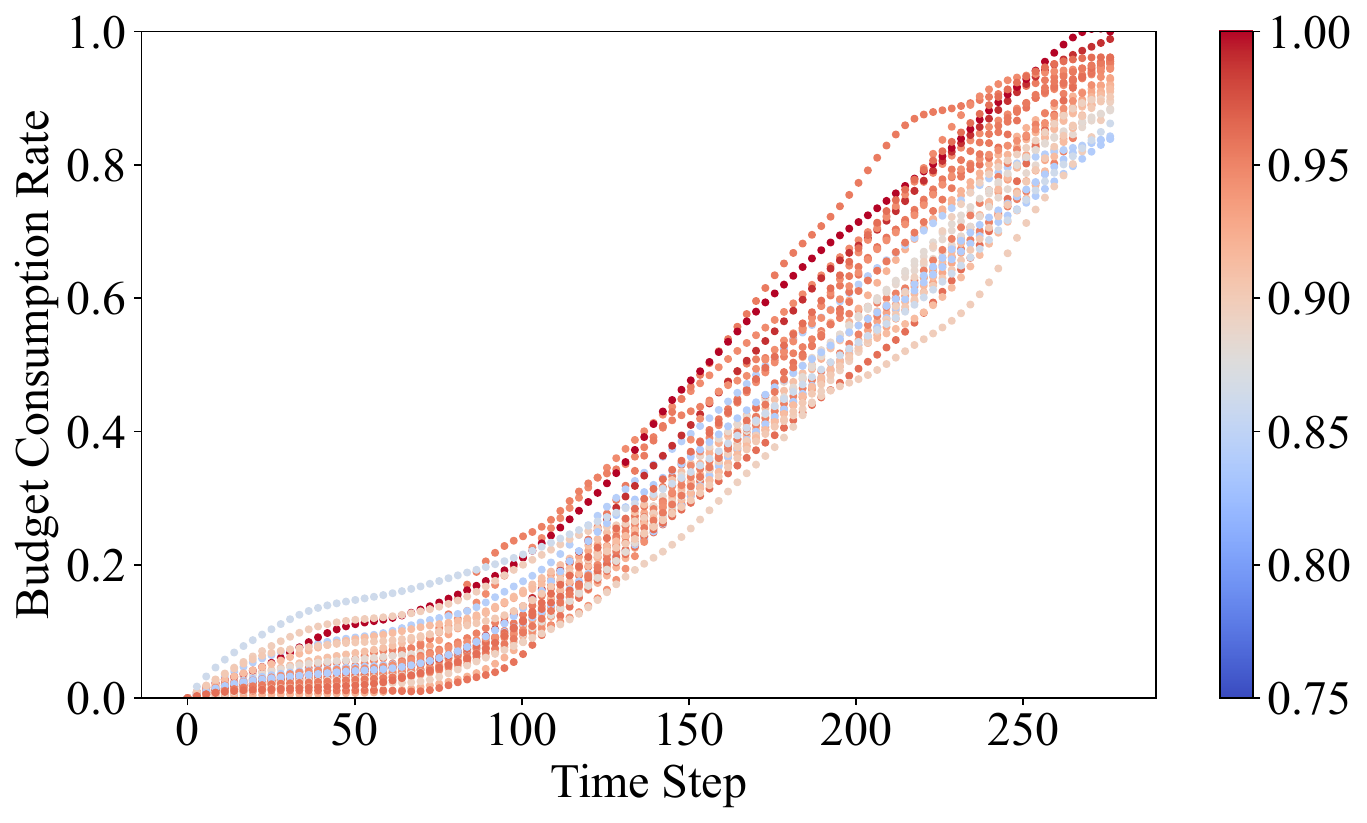}
		\caption{AHBid}
		\label{fig:cost_AHBid}
	\end{subfigure}
  \caption{State transition in one episode.}
	\label{fig:day_cost_transition}
\end{figure}

\begin{figure}[t!]
	\centering
	\begin{subfigure}{0.49\linewidth}
		\centering
		\includegraphics[width=1.0\linewidth]{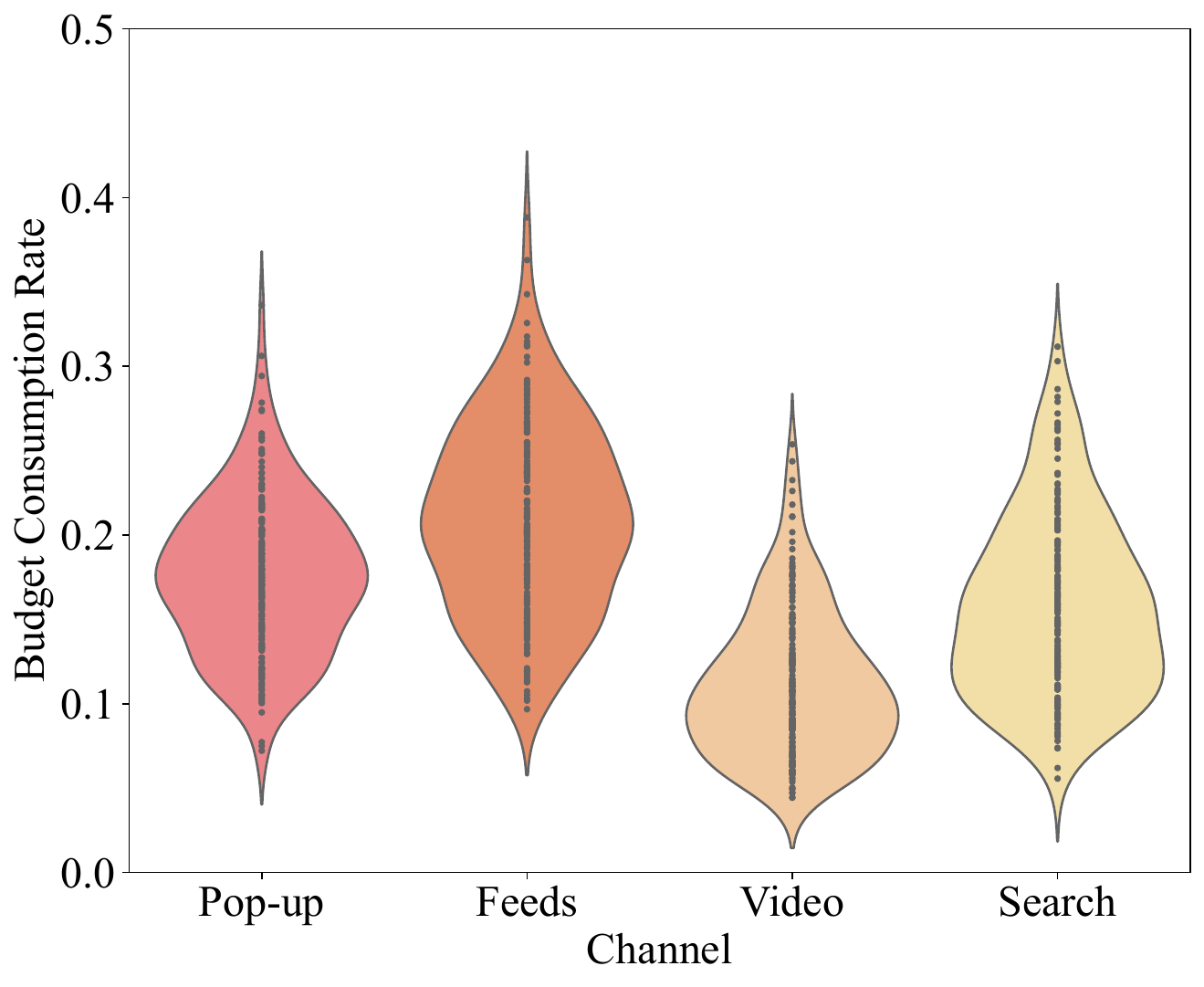}
		\caption{HiBid}
		\label{fig:channel_cost_hibid}
	\end{subfigure}
	\centering
	\begin{subfigure}{0.49\linewidth}
		\centering
		\includegraphics[width=1.0\linewidth]{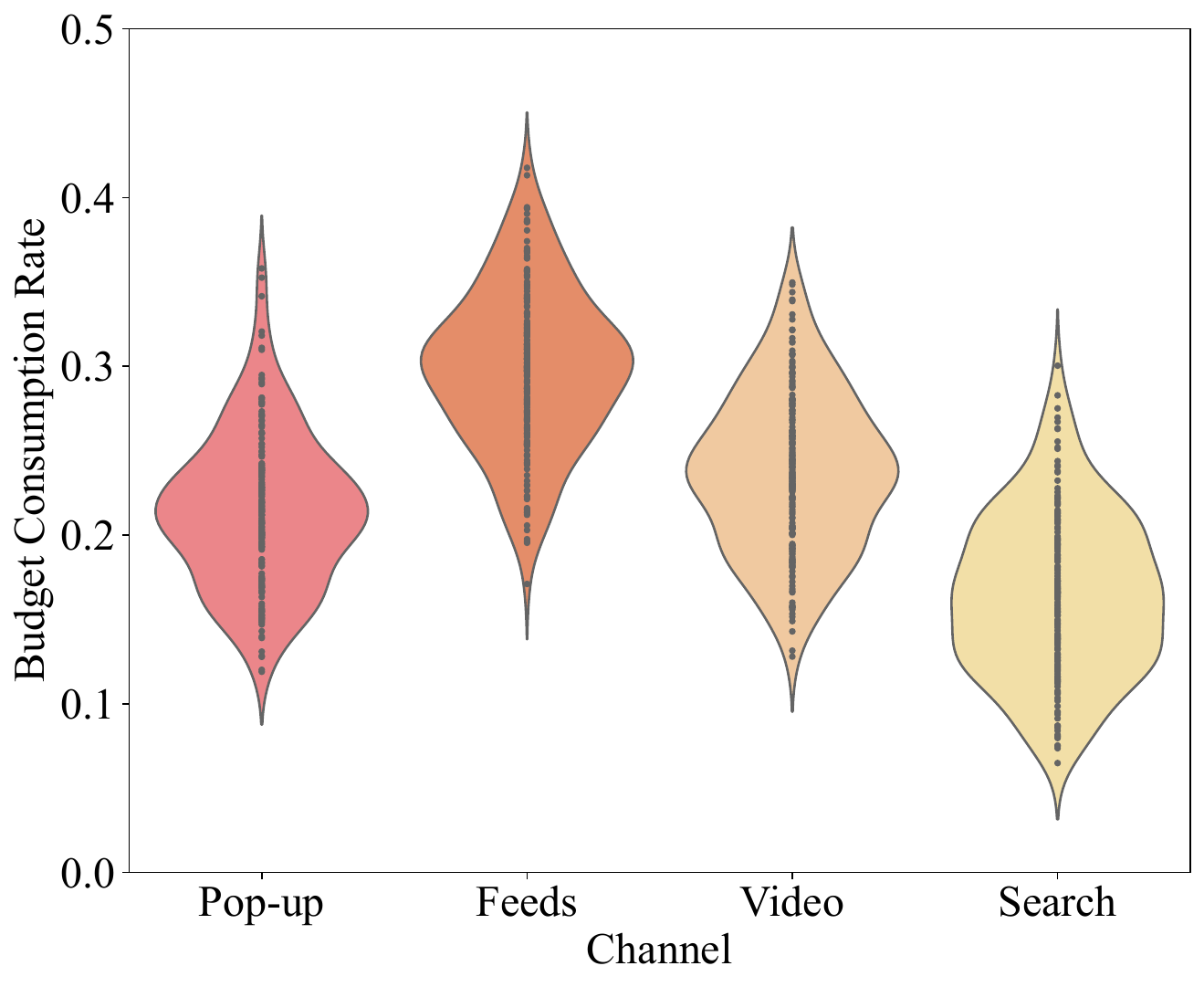}
		\caption{AHBid}
		\label{fig:channel_cost_AHBid}
	\end{subfigure}
  \caption{State distribution across channels.}
	\label{fig:cost_distribution}
\end{figure}

We conduct a comparative evaluation of state transitions and budget consumption distributions between AHBid and the baseline model HiBid. The state transition analysis, illustrated in Figure \ref{fig:day_cost_transition}, depicts the budget consumption rate across successive time steps within an episode. Results indicate that AHBid achieves a higher budget consumption rate than HiBid, reflecting more efficient expenditure pacing throughout the campaign. This enhancement is attributed to AHBid's ability to generate trajectories that optimize budget utilization for improved conversion outcomes.

Additionally, Figure \ref{fig:cost_distribution} presents a channel-wise comparison of budget consumption rates across four channels: Pop-up, Video, Feeds, and Search. Although consumption patterns vary by channel, AHBid consistently achieves improved budget utilization through adaptive reallocation of budgets and dynamic CPC adjustments. These results highlight AHBid's competitive strength, particularly its capacity for fine-grained, channel-level optimization.

\subsection{Online A/B Tests}

\begin{table}
  \caption{Online A/B test results.}
  \label{tab:online_AB_result}
  \begin{tabular}{cccccc}
    \toprule
    Model Type & \#campaign & budget & $R$ & $CSR$ & $BCR$ \\
    \midrule
    HiBid & 2360 & 3342847 & 417268 & 0.836 & 0.897 \\
    AHBid & 2360 & 3342847 & 473892 & 0.904 & 0.935 \\
    \midrule
    Diff &  - & - & 13.57\% & 4.13\% & 5.06\% \\
  \bottomrule
\end{tabular}
\end{table}

To further validate the superiority of AHBid, we deployed the model in a live advertising environment and conducted a two-week A/B test from August 1 to August 14, 2025. This online experiment compared AHBid against the hierarchical baseline model, HiBid. The results, summarized in Table \ref{tab:online_AB_result}, demonstrate that AHBid significantly outperforms HiBid, achieving a 13.57\% increase in return and a 4.13\% improvement in constraint satisfaction rate. These results confirm the practical efficacy and competitive advantage of AHBid in real-world advertising scenarios.
In terms of computational efficiency, the planner generates goal trajectories within 0.04 seconds when accelerated by GPU, which aligns well with the latency tolerances typical in advertising systems.

\section{Conclusion}
\label{sec:conclusion}

In this work, we propose AHBid, an adaptive hierarchical bidding framework designed to enhance auto-bidding performance in cross-channel advertising environments through the integration of generative planning and real-time control. The framework employs a diffusion-based planner that effectively captures temporal dependencies and historical context, enabling robust adaptation to volatile advertising conditions. To ensure constraint compliance and improve responsiveness to environmental dynamics, we introduce a constraint enforcement mechanism coupled with a trajectory refinement strategy. Additionally, a control-based bidding algorithm synergistically integrates historical knowledge with real-time information to enhance operational adaptability and efficacy.
Extensive experimental results and online deployment validate that AHBid significantly enhances both adaptability and performance of auto-bidding strategies, providing an effective solution for optimizing advertising campaigns in complex cross-channel environments.

\bibliographystyle{ACM-Reference-Format}
\balance

\appendix

\section{Appendix}

\subsection{Related Work}
\label{sec:related_work}

\subsubsection{Bidding Strategies}

Bid optimization in online advertising is fundamentally a sequential decision-making problem \cite{HeCWPTYXZ21}. Control-based methodologies \cite{DBLP:conf/kdd/YangLWWTXG19} along with classical Proportional-Integral-Derivative (PID) controllers \cite{bennett1993development} harness online feedback to dynamically adjust bids in real time. An alternative formulation of the auto-bidding problem is MDP and can be tackled using RL techniques. Noteworthy studies, including \cite{wu2018budget, DBLP:conf/kdd/zhao,HeCWPTYXZ21, DBLP:conf/kdd/GuanWCL0LXQXZ21, DBLP:conf/cikm/YuanGXWSZ22}, have effectively employed RL to optimize bidding policies under various constraints. More recently, \cite{DBLP:conf/nips/MouHBXYXZ22} proposed an iterative offline RL framework designed to mitigate the sim2real gap, while \cite{DBLP:conf/kdd/GuoHZWYXZZ24} introduced a conditional diffusion model aimed at addressing MDP-based bidding challenges. Despite their efficacy, these methods primarily focus on single-channel bid optimization and lack scalability to cross-channel scenarios due to their inability to dynamically adjust budget and constraint allocation strategies across channels.

\subsubsection{Multi-Channel Bidding}

In the field of multi-channel bidding, hierarchical frameworks combining a high-level planner and a low-level auto-bidder are commonly employed to allocate budget across advertising channels \cite{deng2023multi}, a problem also referred to as multi-platform bidding \cite{DBLP:conf/www/AvadhanulaCLSS21} or cross-channel campaign optimization \cite{DBLP:conf/kdd/ZhangZGYYL12}. Prior work has explored empirical methods using RL for high-level budget allocation \cite{DBLP:conf/icpr/WangLH18, DBLP:conf/cikm/XiaoGJLCZY19}, derived closed-form solutions based on fitted effectiveness functions for dynamic budget allocation \cite{DBLP:journals/eor/LuzonPK22}, and proposed joint learning of allocation and bidding policies via offline RL \cite{DBLP:journals/ai/NuaraTGR22, DBLP:journals/tc/WangTLMZDSXWW24}. 
Unlike these methods, which primarily rely on current state information for decision-making, AHBid explicitly captures temporal dependencies and observational patterns in historical interaction sequences, providing an adaptive solution for volatile cross-channel bidding environments.

\subsection{Training and Inference of AHBid}
\label{appendix:training_inference_procedure}

\begin{algorithm}[t!]
  \caption{Training of the generative planner}
  \label{algo.training}
  \LinesNumbered
  \KwIn{Randomly initialized $\theta$, trajectory datasets $\mathcal{D}$, condition drop rate $p_u$}
  \KwOut{Parameter $\theta$ of the planner.}
  \Repeat{converged}{
    Sample a batch of trajectories $\mathcal{B} \in \mathcal{D}$ \\
    \ForAll{$\bm{\tau} \in \mathcal{B}$}{
      Perform $\bm{y}(\bm{\tau}) \leftarrow \varnothing$ with probability $p_u$ \\
      Sample $k \sim \text{Uniform} (1, K), \bm{\epsilon} \in \mathcal{N}(0, 1)$ \\
      Calculate $\bm{\tau}_k$ according to Eq. (\ref{eq:forward}) \\ 
      Calculate estimated noise $\hat{\bm{\epsilon}}$ via Eq. (\ref{eq:noise_estimation}) \\
      Calculate $\mathcal{L}_{final}$ via Eq. (\ref{eq:loss_function_hybrid}) \\
      Perform gradient descent to optimize $\theta$ \\
    }
  }
\end{algorithm}

\begin{algorithm}[t!]
  \caption{Bid generation with AHBid}
  \label{algo.inference}
  \LinesNumbered
  \KwIn{The planner $\bm{\epsilon}_\theta$, guidance scale $\omega$, condition $\bm{y}^*(\bm{\tau})$, historical model $\Lambda_{h}$, real-time model $\Lambda_{r}$}
  \KwOut{Bid $b$}
  \ForEach{m = 1, \ldots, M}{
    Get preceding stage states $\bm{s}_{1:m}$\\
    Sample $\bm{\epsilon} \in \mathcal{N}(0, 1)$ \\
    Calculate $\widetilde{\bm{\tau}}_K \sim \mathcal{N}\left(0, \beta_{K}\bm{\epsilon}\right)$\\
    \ForEach{$k = K, \ldots, 1$}{  
      $\widetilde{\bm{\tau}}_k[:m] \leftarrow \bm{s}_{1:m}$ \\
      Estimate noise $\hat{\bm{\epsilon}}_k$ via Eq. (\ref{eq:noise_estimation}) \\
      Calculate $(\bm{\mu}_{k-1}, \bm{\Sigma}_{k-1})$ via Eq. (\ref{eq:parameter_mu}) and Eq. (\ref{eq:parameter_sigma}) \\
      Sample $\bm{\tau}_{k-1} \sim \mathcal{N}\left(\bm{\mu}_{k-1}, \bm{\Sigma}_{k-1}\right)$ \\
    }
    \While{stage not end}{
      Receive impression $o_{i, j}$ \\
      Calculate $b_{i, j}$ via Eq. (\ref{eq:optimal_bidding_solution}) \\
      Update $\Lambda_{r}(t)$ and $\lambda_0, \lambda_1$ via Eq. (\ref{eq:model_combination}) at periodic intervals \\
    }
  }
\end{algorithm}

During training, we employ the well-established Denoising Diffusion Probabilistic Model (DDPM) \cite{DBLP:conf/nips/HoJA20} framework to train the planner. In each iteration, a set of trajectories $\bm{\tau}$ and their associated properties $\bm{y}(\bm{\tau})$ are sampled. With probability $p_u$, the condition $\bm{y}(\bm{\tau})$ is replaced with a dummy value $\varnothing$. The noised trajectory $\bm{\tau}_{k}$ is then obtained using Eq. (\ref{eq:forward}), and the corresponding noise $\hat{\bm{\epsilon}}$ is estimated using Eq. (\ref{eq:noise_estimation}). The planner is updated with the loss computed according to the hybrid objective defined in Eq. (\ref{eq:loss_function_hybrid}). The complete training procedure for AHBid is summarized in Algorithm (\ref{algo.training}). For the low-level bidder, the optimization process involves directly searching for the optimal parameters of the historical model. This procedure is substantially more efficient than training the planner, owing to the fact that the historical model comprises only two parameters.

During inference, the planner makes decisions exclusively at the beginning of each stage, resulting in a total of $M$ decisions per episode. In contrast, the bidder determines bids for every individual incoming impression. The computational complexity of the planner for a single episode is $O(MKH)$, where $M$, $K$, and $H$ represent the number of stages, diffusion steps, and hidden units in the model, respectively. The complete inference procedure of AHBid is detailed in Algorithm \ref{algo.inference}.

\subsection{Notations}

\begin{table}[t!]
  \caption{Important notations in this work}
  \label{tab:notation}
  \begin{tabular}{cp{6cm}}
    \toprule
    Notation&Explanation\\
    \midrule
    $J$ & The number of channels\\
    $I_j$ & The number of impressions in channel $j$\\
    $o_{i, j}$ & The $i$-th impression in channel $j$\\
    $b_{i, j}$ & The bid price for impression $o_{i, j}$ \\
    $v_{i, j}$ & The value of impression $o_{i, j}$ \\
    $B$, $CPC$ & The budget and CPC constraint \\
    $g_{j, m}$ & The goal of stage $m$ in channel $j$ \\
    $s_{j, m}$ & The historical state of stage $m$ in channel $j$ \\
    $\bm{\tau}$ & The original goal trajectory \\
    $\bm{\tau}_k$ & The noised goal trajectory \\
    $\widetilde{\bm{\tau}}_k$ & The estimated goal trajectory \\
    $\bm{y}(\bm{\tau})$ & The condition associated with $\bm{\tau}$ \\
    $R(\bm{\tau})$ & The return associated with $\bm{\tau}$ \\
    $\hat{\bm{\epsilon}}_k$ & The estimated noise\\
    $\bm{\epsilon}_{\theta}$ & The planner's diffusion model \\
    $\Lambda_h$, $\Lambda_r$ & The historical bidder and the real-time bidder \\
  \bottomrule
\end{tabular}
\end{table}

\subsection{Ablation Study}
\label{subsec:ablation_study}

\begin{table}[t!]
  \caption{The ablation study of the bidding model.}
  \label{tab:study_of_combine_model}
  \renewcommand{\arraystretch}{0.93}
  \begin{tabular}{cccc}
    \toprule
    Model Type & $R$ & $CSR$ & $BCR$\\
    \midrule
    Inverse Dynamic Model & 242.35 & 0.796 & 0.821 \\
    Historical Model & \underline{283.14} & 0.814 & \underline{0.912} \\
    Real-time Model & 267.43 & \underline{0.831} & 0.887 \\
    \textbf{Combined Model} & \textbf{298.59} & \textbf{0.907} & \textbf{0.936} \\
    \midrule
    Improvement & 5.45\% & 9.14\% & 2.63\% \\
  \bottomrule
\end{tabular}
\end{table}

To evaluate the impact of the adaptive bidding model, we conducted an ablation study comparing AHBid's performance when its bidding module is replaced with: (1) an inverse dynamic model, (2) a standalone historical model, and (3) a standalone real-time model. The inverse dynamic model was implemented as a two-layer MLP with 512 hidden units and ReLU activations.

Results in Table \ref{tab:study_of_combine_model} show that both the historical and real-time models significantly outperform the inverse dynamic model, which struggles to adapt to highly dynamic advertising conditions. The historical model exhibits robustness through MPC-based control, while the real-time model leverages immediate feedback. The combined model achieves the highest performance, with a 5.45\% improvement in return and a 9.14\% increase in constraint satisfaction rate, highlighting the benefit of integrating historical and real-time information for adaptive bidding.

\end{document}